\documentclass[sigconf]{acmart}

\AtBeginDocument{%
  \providecommand\BibTeX{{%
    \normalfont B\kern-0.5em{\scshape i\kern-0.25em b}\kern-0.8em\TeX}}}

\usepackage{multirow} 

\renewcommand\footnotetextcopyrightpermission[1]{} 

\usepackage{fancyhdr}
\makeatletter
\def\@maketitle{%
  \newpage
  \null
  \vskip 2em%
  \begin{center}%
    {\LARGE \@title \par}%
    \vskip 1.5em%
    {\large
      \lineskip .5em%
      \begin{tabular}[t]{c}%
        \@author
      \end{tabular}\par}%
    \vskip 1em%
    {\large \@date}%
  \end{center}%
  \par
  \vskip 2.5em}
\makeatother

\begin{document}

\title{GUI Element Detection Using SOTA YOLO Deep Learning Models}

\author{Seyed Shayan Daneshvar}
\email{daneshvs@myumanitoba.ca}
\orcid{0000-0002-3463-7873}
\affiliation{%
  \institution{University of Manitoba}
  \city{Winnipeg}
  \state{Manitoba}
  \country{Canada}
}

\author{Shaowei Wang}
\email{Shaowei.Wang@umanitoba.ca}
\orcid{0000−0003−3823−1771}
\affiliation{%
  \institution{University of Manitoba}
  \city{Winnipeg}
  \state{Manitoba}
  \country{Canada}
}


\renewcommand{\shortauthors}{S. Daneshvar and S. Wang}

\begin{abstract}
  Detection of Graphical User Interface (GUI) elements is a crucial task for automatic code generation from images and sketches, GUI testing, and GUI search. Recent studies have leveraged both old-fashioned and modern computer vision (CV) techniques. Old-fashioned methods utilize classic image processing algorithms (e.g. edge detection and contour detection) and modern methods use mature deep learning solutions for general object detection tasks. GUI element detection, however, is a domain-specific case of object detection, in which objects overlap more often, and are located very close to each other, plus the number of object classes is considerably lower, yet there are more objects in the images compared to natural images. Hence, the studies that have been carried out on comparing various object detection models, might not apply to GUI element detection. In this study, we evaluate the performance of the four most recent successful YOLO models for general object detection tasks on GUI element detection and investigate their accuracy performance in detecting various GUI elements.  
\end{abstract}




\received{10 April 2023}

\maketitle

\section{Introduction}

Users interact with software applications through GUI elements such as widgets, images, texts, etc. Many software engineering tasks, such as code generation \cite{image2emmet, UID2GUI2018, CodeGenEncDec2022, REMAUI, MLPrototypeMobile2020}, GUI testing and automation \cite{GuiTestWidget2019, AutoReportMobileDesignViolation2018}, GUI search \cite{Rico, VINS}, advanced GUI interaction support \cite{WakenSV}, and UI generation \cite{GUIGAN} require GUI element detection as the initial step. GUI element detection and recognition techniques are either based on instrumentation \cite{ActivitySpace4Instr} or pixels, where the former requires run-time infrastructures that expose GUI elements' information or accessibility APIs. The latter, however, use the image of GUIs for analysis and are more general and easier to use.

In all pixel-based GUI recognition methods, object detection techniques from computer vision (CV) and deep learning are used. It is worth noting that general object detection in computer vision is mostly applied in natural images where objects do not overlap very often and there are not many interesting objects to detect in natural scenes. This is not the case with GUI element detection as elements can be placed on top of or very close to each other in a very small space, such as the screen of a mobile phone.

Object detection in computer vision refers to the detection of object instances of a certain class (such as apple, orange, or banana) in images and videos. Most object detection techniques consist of two sub-tasks: {\itshape region proposal} or {\itshape detection}, in which the location of the bounding box containing the object is predicted, and {\itshape region classification}, which resolves the class of the object in the bounding box. 

Deep learning-based object detection \cite{ObjDetSurveyAnchor2020, yolov4, yolov5, yolov6r3, Yolo7,yolov8} can be classified into two broad categories: anchor-based and anchor-free methods. Anchor-based methods can be divided into two categories of two-stage methods, which are based on region detection or proposal, and single-stage methods which are based on regression. Two-stage anchor-based methods (such as Faster R-CNN \cite{FasterRCNN}) consist of two stages, extraction of region proposals by an algorithm, and using a CNN for classification and location refinement of the detected bounding box. One-stage anchor-based methods such as the YOLO series, do not generate region proposals and directly map the location of the bounding box into the regression of the bounding box points. Anchor-free methods are divided into two categories as well: key-point-based methods such as Centernet \cite{Centernet}, where the bounding box is described by location key points, and dense-predication methods such as FCOS \cite{FCOS}, which are based on the idea of semantic segmentation in computer vision, where the object's center and its boundary distance is predicted.

YOLO is referred to as a series of fast and highly accurate one-stage object detection algorithms based on CNNs that started with the works of Redmon et al. \cite{YOLO} and researchers continued to improve and enhance the original version by using more advanced architectures, techniques, and loss functions \cite{YOLOv2, YOLO3, yolov4, yolov5, yolov6r3, Yolo7, yolov8}. YOLOv4 reorganized the architecture into 3 parts, namely backbone, neck, and head; later models followed the same architecture and introduced changes in different parts. The backbone is a CNN with the responsibility of feature extraction and usually uses a pre-trained network such as Resnet \cite{resnet}. The head produces the output, which includes the predicted bounding boxes, class probabilities, and objectness scores. The neck is an optional part of the architecture which connects the backbone to the head, which can extract more features, reduce dimensionality, or carry out other operations to improve the performance even further. 

All YOLO architectures have been anchor-based except for YOLOv8 \cite{yolov8} which uses an anchor-free method. Latest versions of YOLO are the seventh version of YOLOv5 \cite{yolov5}, the third version of YOLOv6 \cite{yolov6r3}, YOLOv7 \cite{Yolo7}, and YOLOv8 \cite{yolov8}. It is important to note that not all YOLOs follow the same naming pattern, and some older methods use different names, e.g. YOLOX \cite{yolox}, YOLOR \cite{yolor}, and PP-YOLOE \cite{yoloe}. YOLOv5 and YOLOv8 have 5 pre-trained models with different model sizes for different platforms, namely N, S, M, L, and X. YOLOv6 similarly has all of these models except for the X, which is the largest model. The N models, which stand for nano, are designed to be small enough to be used in edge devices with limited resources. The S and M variants are for mid-range and high-end desktop and laptop GPUs and systems respectively, and the L and X models are suitable for cloud GPUs. YOLOv7 has 3 basic models for edge, desktop or normal, and cloud GPUs, namely YOLOv7-tiny, YOLOv7, and YOLOv7W6. 

In most works on GUI element detection, a bottom-up approach is adopted where primitive features, shapes, and regions (e.g. edges, corners, and contours) are aggregated into objects. There have been works that utilized old-fashioned \cite{REMAUI, MLPrototypeMobile2020} and modern methods\cite{CodeGenEncDec2022, GUIGAN, image2emmet}, or both\cite{UIED}. Methods based on deep learning, use convolutional neural networks (CNN) with large image datasets for feature extraction and aggregation into objects. Some works utilized both old-fashioned and deep learning methods to get the best results. In contrast, some works\cite{UIED} utilized a top-down approach and leveraged both old-fashioned methods and deep learning to get better results. 

The recent studies on GUI element detection\cite{UIED, UIEDprior2020, MLPrototypeMobile2020} have tested various famous deep-learning object detection techniques. Chen et al. \cite{UIEDprior2020} compared various models' performance on previous old-fashioned methods as two deep learning-based methods, namely YOLOv3 and Centernet, plus they came up with a hybrid method combining old-fashioned CV and deep learning methods to detect non-text elements; they used an already available text detector to detect and localize GUI elements. Xie et al. \cite{UIED} implemented the methods compared by Chen et al. as well as their suggested method, and made an interactive tool for the users, with the ability to change and edit the detection results. Xu et al. \cite{image2emmet} collected web images alongside their HTML-CSS code, then utilized region-based object detection deep learning models, and finally generated corresponding code for the components via a deep learning model, combining CNN and LSTM. Bunian et al. \cite{VINS} introduced a visual search framework for UI designs; they gathered a dataset containing 4800 UI images, 257 of which are abstract wireframes, and the remaining 4543 images are GUI images; they used their collected dataset of GUI images (without wireframes) to train a Single Shot Multibox Detector (SSD) for GUI element detection, and then used the detector to find similar UI designs.

Recently, YOLOv8, YOLOv6 v3.0 (YOLOv6R3), YOLOv7, and YOLOv5 v7 (YOLOv5R7) came out, and they all seem to perform relatively similar to each other for general object detection tasks \cite{yolov6r3}. The goal of this study is to compare these recent models for GUI element detection, especially YOLOv5R7 as all previous comparisons in general object detection used the sixth version which performs slightly worse than the latest model. We will also investigate which GUI elements are easier, and which are harder to correctly detect and classify.

For evaluation, we will report the precision, recall, F1 score, mean average precision (mAP@0.5), and AP@[.5:.05:.95]. All of these metrics except for the last one will be calculated on measurements with an IoU of over 0.5. IoU is the intersection area over the union area of the detected bounding box and ground truth. Precision in object detection is the number of correct detections (True Positives) divided by all detections. Recall is the number of true positives divided by the number of all ground truths. F1 score is used as a primary evaluation metric as it is interpreted as the harmonic mean of precision and recall. Mean Average Precision with an IoU threshold of 0.5 (mAP@0.5) is the mean of AP@0.5 for all classes, where AP@0.5 is the area under the precision-recall curve measured at IoU > 0.5. Finally, AP@[.5:.05:.95] is the same as mAP but it is averaged for IoUs between 0.5 and 0.95 with a step of 0.05. This last metric is used less often but gives good intuition about the precision of the model when the IoU threshold increases, i.e. a higher localization accuracy is required.


Our study makes the following contributions:
\begin{itemize}
\item We implement and compare the 4 most recent state-of-the-art models from YOLO families for GUI element detection.
\item We investigate which GUI elements are harder and which are easier to detect for each model, and for all models.
\item Finally, we verify whether the models follow the same order in terms of performance for GUI element detection.
\end{itemize}

\section{Empirical Study}
In this study, to answer the previously mentioned unanswered question, we carry out the first empirical study of using the latest YOLO models for GUI element detection. Our study involves the use of the VINS dataset \cite{VINS} with 4543 images of Android and IOS mobile applications for GUI element detection with the latest YOLO object detection models, namely YOLOv5R7, YOLOv6R3, YOLOv7, YOLOv8. 
we ask and answer the following research questions:

\subsection{Research Questions}
In this study, we focus on the following research questions to find out which model should be preferred for GUI element detection, which GUI elements are trickier to detect and check whether GUI element detection results verify the results of previous studies on general datasets such as MS COCO \cite{coco}.
\begin{itemize}

\item \textbf{RQ 1 Accuracy Performance:} 
How precisely can these models detect and classify the GUI elements? i.e. How high will their recall, precision, f1, and mAP scores be? Will these models beat the work of Bunian et al \cite{VINS}?
\item \textbf{RQ 2 Element Detection Difficulty:} 
Which elements are harder to detect for each model, and which elements are universally hard for all models? Which elements are easier, and which are hard for all models to detect and classify? 
\item \textbf{RQ 3 Verification:} 
Do the models' performances for GUI element detection follow the same order as their performances on general datasets? i.e. On general datasets, YOLOv7 achieves the best results in terms of accuracy compared to the other models that we used; Do we get similar results or not?
\end{itemize}

\subsection{Experiment Setup}
\subsubsection{Dataset}
In this study, we will use the VINS dataset provided by Bunian et al. \cite{VINS} and for the results to be comparable to theirs we do not use the abstract wireframes available in the dataset. We split the 4543 images, into 3 sets of train, validation, and test, having the ratio of 80\%, 10\%, and 10\% respectively in a random fashion. The dataset has images with annotations containing 18 classes of elements, namely BackgroundImage, CheckedTextView, Icon, EditText, Image, Text, Text Button, Drawer, PageIndicator, UpperTaskBar, Modal, Switch, Spinner, Card, Multi-tab, Toolbar, Bottom-Navigation, and Remember. Same as Bunian et al. we removed the last 6 labels in the images' annotations, as they are very scarce in the dataset. Moreover, Bunian et al. used alternative names for some of the labels, namely they used InputField, CheckedView, SlidingMenu, and Pop-Up Window to refer to EditText, CheckedTextView, Drawer, and Modal respectively. We use the original naming of elements to avoid confusion.
\begin{figure}[h!]
  \centering
  \includegraphics[scale=0.75]{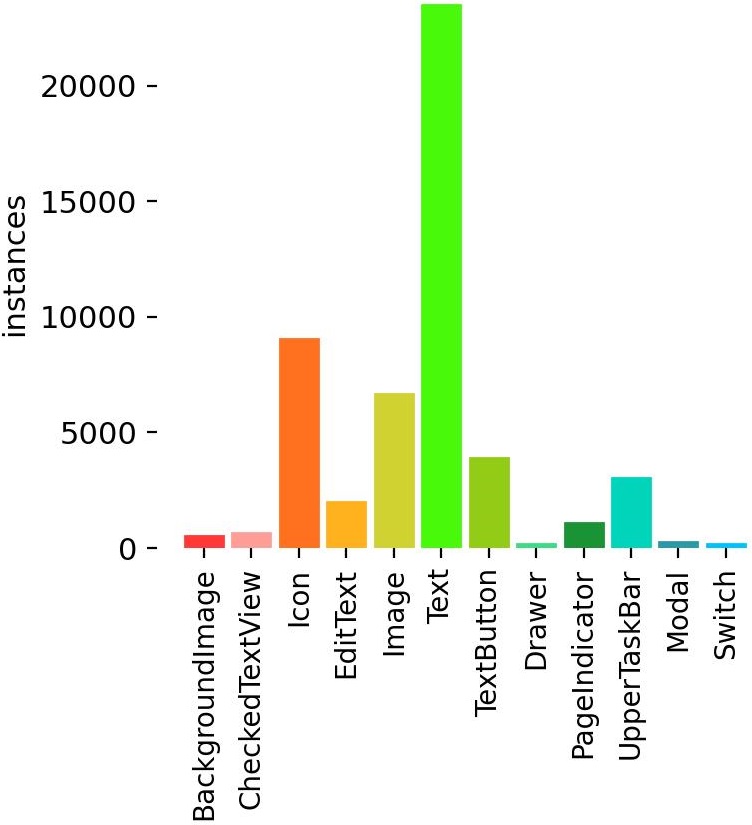}
  \caption{Distribution of selected labels in the dataset.}
  \Description{Text is the dominating class in the labels, and Drawer is the least common.}
  \label{fig:ld}
\end{figure}

The original dataset's annotations for objects are stored in XMLs using Pascal VOC format, where for every object in the image, there exists a label and the two coordinates that form the object's bounding box. This is incompatible with the required input for the YOLO models as the required input for these YOLOs is a TXT file containing the encoded label, the normalized coordinate of the bounding box's center, and the normalized width and height of the bounding box, each number separated from other with a single space character. The encoded label is a constant number assigned to each class of elements, and normalized width and height are easily calculated by subtracting the start and end point coordinates of the bounding box and dividing the results by the size of the image. Similarly, the normalized bounding box centers are easily calculated as the average of bounding box coordinates and divided by the size of the image. Accordingly, we transformed the XMLs to TXT and abandoned irrelevant information available in the XMLs, such as the object's pose, occlusion, and difficulty.

\subsubsection{Model Selection}
Details of the selected models in this study:

\textbf{YOLOv7} YOLOv7 is the oldest model that we are using and it only has one pre-trained variant which is suitable for desktop and laptop GPUs, which is YOLOv7 without any specific tags. Hence, we choose the only available option that is going to be useful for GUI element detection, as GUI element detection will be used in applications such as code generation where code is run on developers' systems, which usually are mid-range systems.

\textbf{YOLOv5R7, YOLOv6R3, and YOLOv8} We selected the small (S) variants of pre-trained variants of these models, which are suitable for low to mid-range desktops and laptops and are smaller than YOLOv7. The reason we did not choose the medium (M) variants is the fact that the S variants give comparable results compared to YOLOv7, and all M variants will beat YOLOv7. It is worth noting that YOLOv7's parameters are on the order of 37 million, and the M variants of YOLOv5R7, YOLOv6R3, and YOLOv8 have around 21, 35, 26 million parameters, and the S variants have around 7, 19, 11 million parameters respectively; meaning that the YOLOv7 variant is actually comparable to the M variants of the other models, in terms of parameters.

\subsubsection{Training Process}
For all of the models, we used the pre-trained weights, which were trained on the MS COCO dataset, and we used the default hyperparameter tuning configuration that comes with each model. All of the models were trained for 300 epochs, in batches of 16 on an RTX 3070 Ti laptop GPU with resized images of size (416,416). All of the models generated their best weights based on the validation dataset before the 268th epoch, and the improvement in the losses of all models slowed down considerably after approximately the 120th epoch. It is worth noting that all of the reported results use the last weight and not the best weight as the best weight is considered best, based on the validation dataset, and in our experience, the last weights achieve better scores on the test set.

\begin{table}[h!]
    \centering
\begin{tabular}{ |p{2.2cm}||p{1cm}|p{0.85cm}|p{0.85cm}|p{0.7cm}| p{0.7cm}|  }
 \hline
 Label & Bunian et al.\cite{VINS} & V5R7s & V6R3s & V7 & V8s\\
 \hline
BackgroundImage & 89.33 & 89.2 & 91.9 & 92.2 & 92.4\\
CheckedTextView & 44.48 & 73.7 & 71.1 & 74.0 & 72.5\\
Icon & 50.50 & 95.4 & 93.9 & 96.4 & 91.4\\
EditText & 78.24 & 97.4 & 97.4 & 98.8 & 96.7\\
Image & 79.24 & 93.1 & 91.0 & 94.2 & 91.2\\
Text & 63.99 & 96.5 & 96.7 & 96.7 & 95.1\\
TextButton & 87.37 & 97.7 & 97.7 & 98.2 & 98.2\\
Drawer & 100 & 99.5 & 99.3 & 99.6 & 99.5\\
Page Indicator & 59.37 & 98.6 & 95.6 & 99.3 & 90.4\\
UpperTaskBar & 90.4 & 99.5 & 96.9 & 73.1 & 99.5\\
Modal & 93.75 & 97.3 & 97.3 & 98.4 & 97.7\\
Switch & 80.00 & 99.3 & 99.5 & 99.6 & 98.0\\
\hline
All (\%) & 76.39 & 94.8 & 94.0 & 93.4 & 93.6\\
 \hline
\end{tabular}

    \caption{Mean average precision at IoU of 0.5 (AP@.5) of the selected methods on the test set, and Bunian et al.\cite{VINS}}
    \label{table:ap}
\end{table}

\subsection{Results - RQ1 Accuracy Performance}

Accuracy results of each model can be seen in detail in tables: \ref{table:v5}, \ref{table:v6}, \ref{table:v7}, and \ref{table:v8}. The summarized table \ref{table:ap} contains the mean average precision of every model at IoU of 0.5 plus the GUI element detection results of Bunian et al \cite{VINS}. As is described in table \ref{table:ap}, all of the models outperform the work of Bunian et al. YOLOv5R7s does 0.13 \% worse than Bunian et al's work in terms of AP@.5 on detecting background images, while others do better; YOLOv7 does 17.3 \% worse than Bunian et al's work in terms of AP@.5 on detecting Upper Taskbars; All of the models perform slightly worse than Bunian et al's work on detecting drawers, yet all of them achieve an AP@.5 of over 99.3 \%. Our selected models outperform Bunian et al's work for all the other classes. 

Overall YOLOv5R7s outperforms Bunian et al's work, YOLOv6R3s, YOLOv7, and YOLOv8s by 18.41\%, 0.8\%, 1.4\%, and 1.2\% in terms of AP@.5 respectively. Hence, we can conclude that YOLOv5R7s is the best of these models if detection at an IoU > 0.5 is sufficient.

IoU > 0.5 is acceptable if elements are placed at a reasonable distance from each other so that the detection result can be fine-tuned with classic computer vision algorithms (Such as Corner SubPixel algorithm \cite{forstner1987fast}) for most GUI elements relatively easily. This, however, might not be the case in mobile GUI images, as elements tend to be placed closer to each other. In such cases, the IoU threshold should be set higher, in which case, the AP@[0.5:0.95] metric gives better insight. In terms of AP[0.5:0.95] YOLOv7 outperforms YOLOv5R7s, YOLOv6R3s, and YOLOv8s by 3.2\%, 3.4\%, and 1.3\%  respectively; this is due to the fact that YOLOv7 has a lot more parameters and this makes it to be more stable in terms of performance as the IoU threshold increases. Hence, it is obvious that the M variant of other models will outperform YOLOv7 as their S variant achieves superior results in terms of AP@.5, and it is likely that YOLOv8m will outperform others' M variants as among the S models, YOLOv8 outperforms others in terms of AP@[0.5:0.95].

\begin{table}[h!]
    \centering
\begin{tabular}{ |p{2.2cm}||p{1.05cm}|p{1cm}|p{1cm}|p{1.75cm}|  }
 \hline
 Label & Precision & Recall & AP@0.5 & AP@[0.5:0.95]\\
 \hline
BackgroundImage & 84.2 & 85.5 & 89.2 & 84.7 \\
CheckedTextView & 95.2 & 59.2 & 73.7 & 52.8 \\
Icon & 94.1 & 86.9 & 95.4 & 75.4 \\
EditText & 96.3 & 95.7 & 97.4 & 86.7\\
Image & 89.8 & 83.6 & 93.1 & 84.2 \\
Text & 96.4 & 95.8 & 96.5 & 83.4 \\
TextButton & 97.1 & 96.1 & 97.7 & 94.9\\
Drawer & 98.2 & 100 & 99.5 & 99.5\\
Page Indicator & 100 & 96.9 & 98.6 & 69.4 \\
UpperTaskBar & 99.7 & 99.6 & 99.5 & 95.1\\
Modal & 93.0 & 96.8 & 97.3 & 92.6\\
Switch & 89.2 & 100 & 99.3 & 88.9\\
\hline
All (\%) & 94.4 & 91.3 & 94.8 & 84.0\\
 \hline

\end{tabular}
    \caption{YOLOv5R7s evaluation results on the test set.}
    \label{table:v5}
\end{table}

\begin{table}[h!]
    \centering
\begin{tabular}{ |p{2.2cm}||p{1.05cm}|p{1cm}|p{1cm}|p{1.75cm}|  }
 \hline
 Label & Precision & Recall & AP@0.5 & AP@[0.5:0.95]\\
 \hline
BackgroundImage & 84.9 & 88.7 & 91.9 & 90.9\\
CheckedTextView & 72.2 & 73.2 & 71.1 & 54.4\\
Icon & 89.6 & 85.8 & 93.9 & 75.1\\
EditText & 94.8 & 94.1 & 97.4 & 85.4\\
Image & 86.8 & 87.3 & 91.0 & 84.5\\
Text & 95.0 & 94.0 & 96.7 & 81.8 \\
TextButton & 95.2 & 96.3 & 97.7 & 95.9\\
Drawer & 88.4 & 100 & 99.3 & 99.3\\
Page Indicator & 98.5 & 82.6 & 95.6 & 69.8\\
UpperTaskBar & 98.8 & 67.1 & 96.9 & 80.3\\
Modal & 85.3 & 96.8 & 97.3 & 95.8\\
Switch & 98.6 & 100 & 99.5 & 92.0\\
\hline
All (\%) & 90.7 & 88.8 & 94.0 & 83.8\\
 \hline

\end{tabular}

    \caption{YOLOv6R3s evaluation results on the test set.}
    \label{table:v6}
\end{table}

\begin{table}[h!]
    \centering
\begin{tabular}{ |p{2.2cm}||p{1.05cm}|p{1cm}|p{1cm}|p{1.75cm}|  }
 \hline
 Label & Precision & Recall & AP@0.5 & AP@[0.5:0.95]\\
 \hline
BackgroundImage & 82.2 & 90.3 & 92.2 & 89.8\\
CheckedTextView & 86.0 & 69.0 & 74.0 & 65.1\\
Icon & 93.7 & 91.9 & 96.4 & 82.9\\
EditText & 96.9 & 95.9 & 98.8 & 93.8\\
Image & 88.3 & 86.5 & 94.2 & 88.5\\
Text & 96.5 & 0.95.5 & 96.7 & 87.6\\
TextButton & 96.7 & 96.5 & 98.2 & 97.4\\
Drawer & 97.5 & 100 & 99.6 & 99.6\\
Page Indicator & 99.3 & 96.6 & 99.3 & 79.2\\
UpperTaskBar & 99.6 & 62.0 & 73.1 & 68.0\\
Modal & 93.6 & 94.2 & 98.4 & 97.6\\
Switch & 89.7 & 100 & 99.6 & 96.7\\
\hline
All (\%) & 93.3 & 89.9 & 93.4 & 87.2\\
 \hline

\end{tabular}

    \caption{YOLOv7 evaluation results on the test set.}
    \label{table:v7}
\end{table}

\begin{table}[h!]
    \centering
\begin{tabular}{ |p{2.2cm}||p{1.05cm}|p{1cm}|p{1cm}|p{1.75cm}|  }
 \hline
 Label & Precision & Recall & AP@0.5 & AP@[0.5:0.95]\\
 \hline
BackgroundImage & 85.6 & 86.5 & 92.4 & 91.0\\
CheckedTextView & 76.4 & 64.8 & 72.5 & 61.5\\
Icon & 89.5 & 81.7 & 91.4 & 74.1\\
EditText & 93.4 & 96.7 & 96.7 & 90.9\\
Image & 85.9 & 82.2 & 91.2 & 85.0\\
Text & 95.7 & 88.6 & 95.1 & 85.2\\
TextButton & 95.0 & 97.4 & 98.2 & 96.9\\
Drawer & 93.2 & 100 & 99.5 & 99.5\\
Page Indicator & 99.1 & 68.7 & 90.4 & 66.0\\
UpperTaskBar & 99.7 & 99.7 & 99.5 & 91.0\\
Modal & 86.6 & 93.5 & 97.7 & 97.2\\
Switch & 98.9 & 97.4 & 98.0 & 92.5\\
\hline
All (\%) & 91.6 & 88.1 & 93.6 & 85.9\\
 \hline

\end{tabular}

    \caption{YOLOv8s evaluation results on the test set.}
    \label{table:v8}
\end{table}


\subsection{Results - RQ2 Element Detection Difficulty}
\begin{figure*}[h!]
    \centering
    (a){\includegraphics[width=0.479\textwidth]{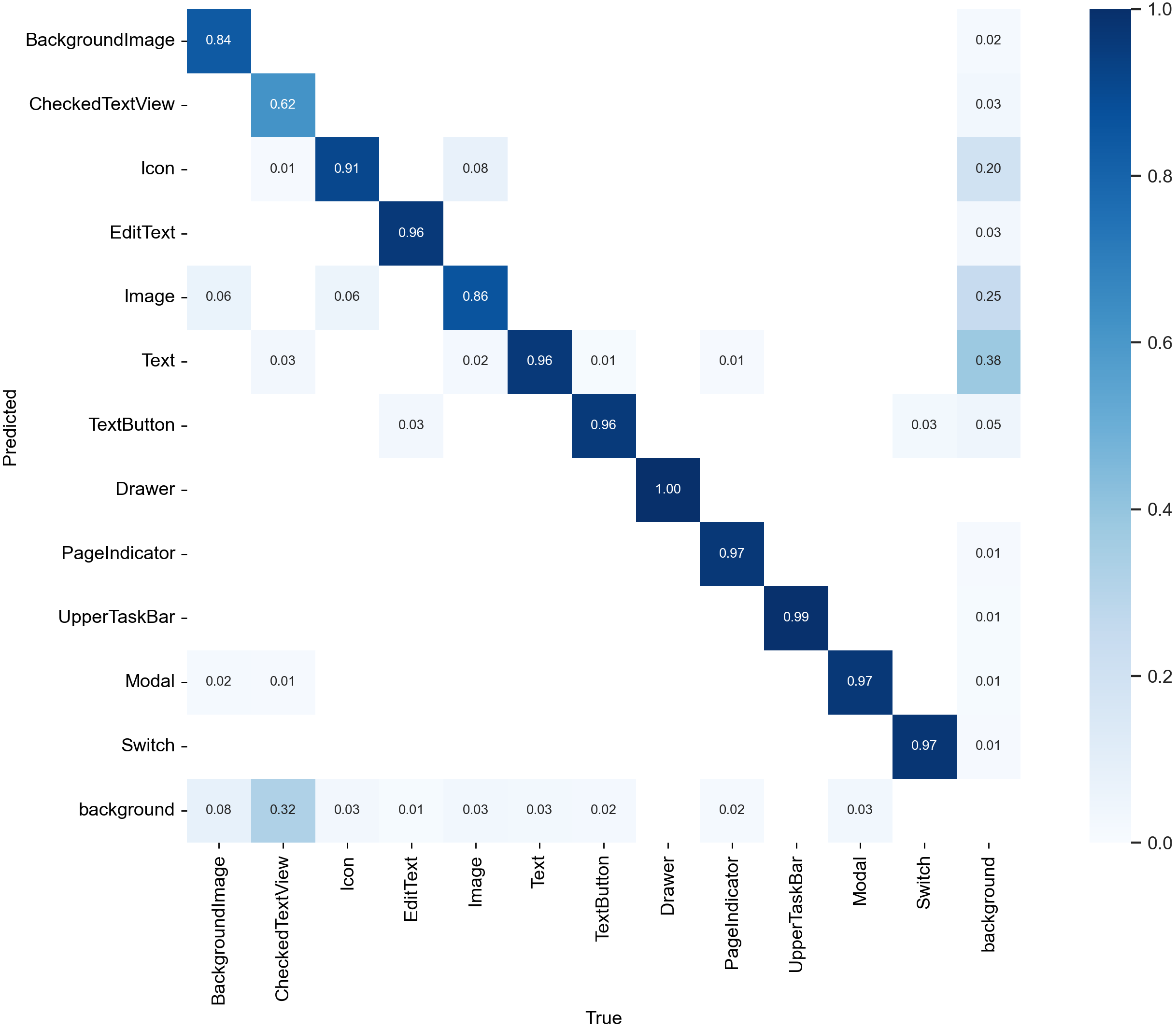}}
    (b){\includegraphics[width=0.48\textwidth]{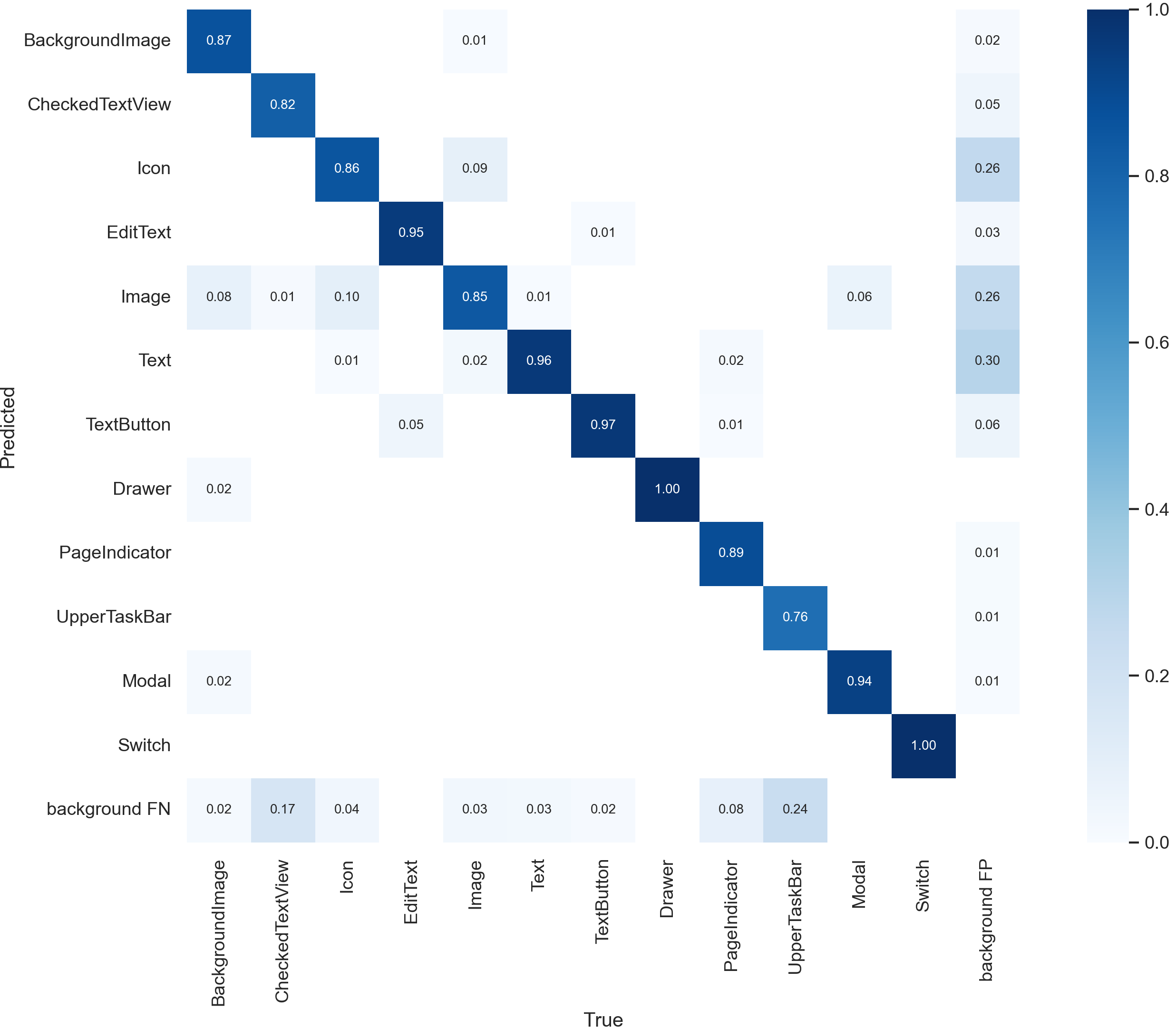}}
    (c){\includegraphics[width=0.48\textwidth]{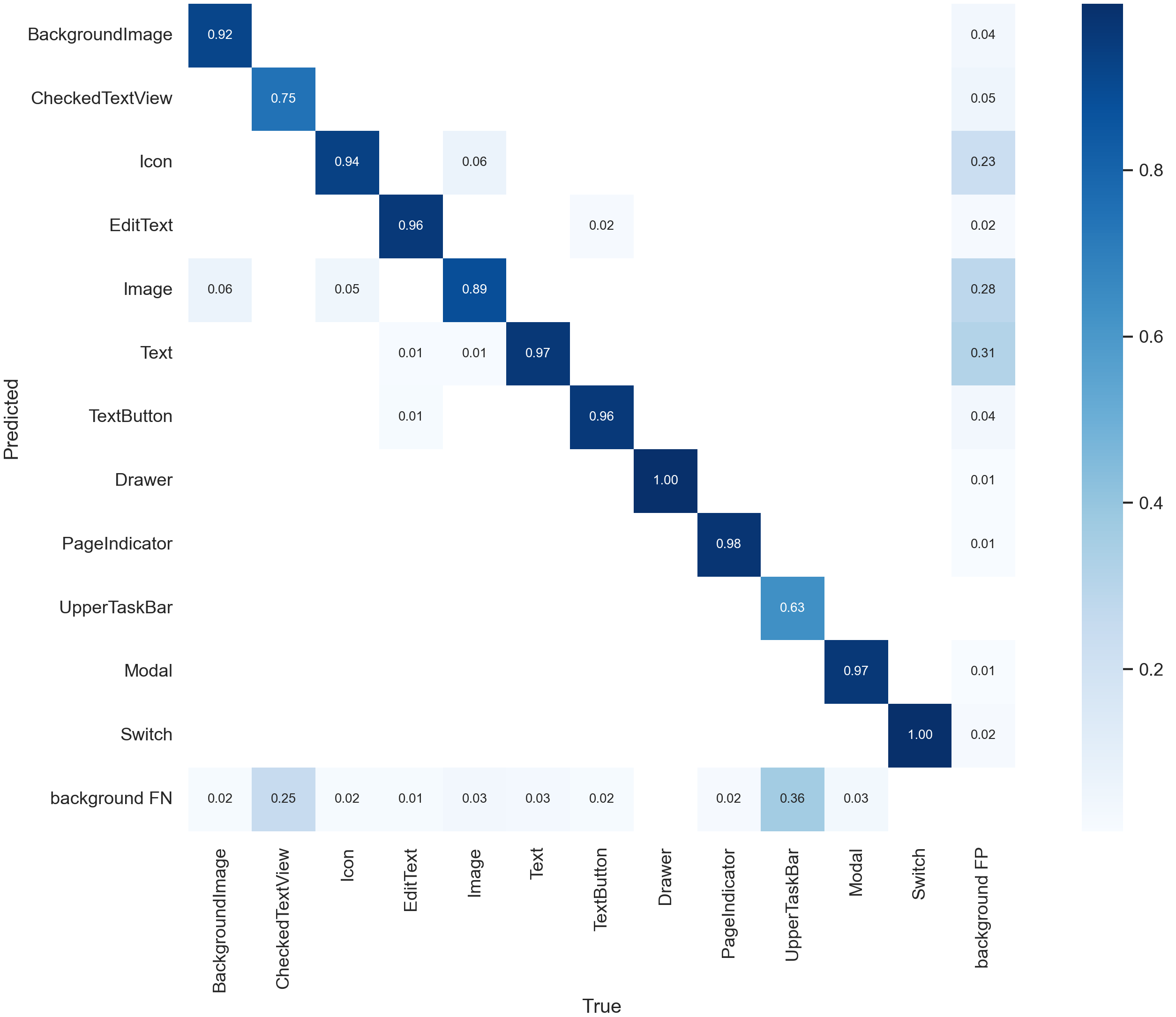}}
    (d){\includegraphics[width=0.48\textwidth]{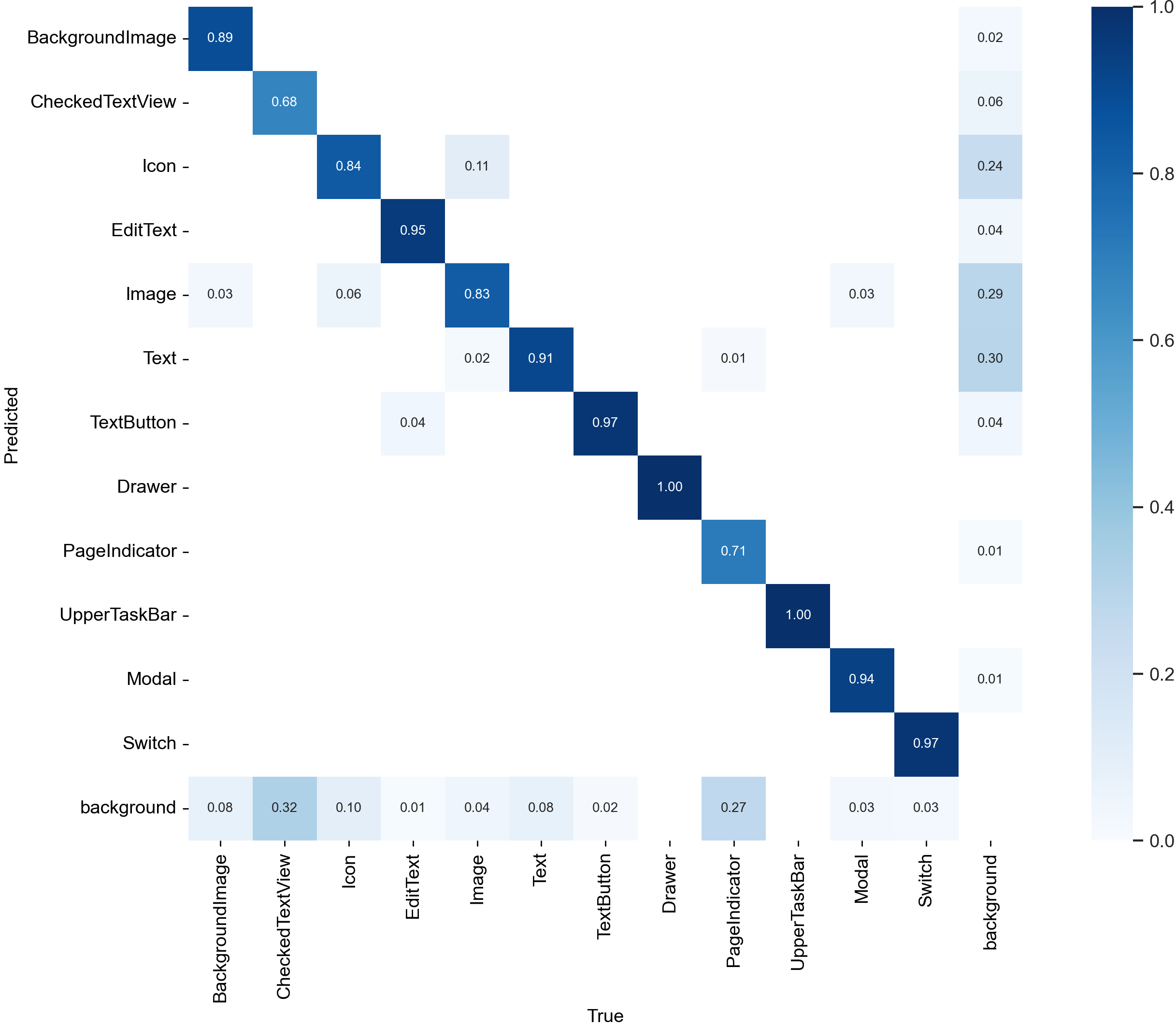}}
    \caption{Confusion matrices on the test set. (a) YOLOv5R7s (b) YOLOv6R3s (c) YOLOv7 (d) YOLOv8s}
    \label{fig:cm}
\end{figure*}

As it is described in figure \ref{fig:cm}, YOLOv6s and YOLOv7 have difficulty with the detection and classification of Upper Taskbars, and all of the models with Checked Text Views. On the other hand, YOLOv5R7s and YOLOv8s perform extremely well for Upper Taskbars, and all the models have minimum to no difficulty with detecting and classifying Drawers and Switches. 

YOLOv8s performs relatively poorly in detecting Page Indicators, while all the other models perform reasonably well in detecting Page Indicators. Page Indicators are the smallest type of element among other elements in the dataset, and they have the highest width-to-height ratio, yet they are more common than 5 other classes of elements and do not vary much in terms of appearance compared to other elements. It also performs worse than others on detecting switches in terms of AP@.5 and it is the only model that does not achieve a recall rate of 100\% on switches and misses switches sometimes. Switches are another element with a high width-to-height ratio, similar to Page Indicators. Hence, it can be concluded that YOLOv8s is more sensitive to the width-to-height ratio than others and performs worse when the width-to-height ratio is too high.

On the other hand, YOLOv8s achieves the best accuracy on detecting Upper Taskbars which do have a high width-to-height ratio, but they are always placed at the top of the image, and have the least variation among all the other elements, thus it can be concluded that YOLOv8s can detect and classify objects with minimum translational invariance, while other models tend to be more agnostic to the location of the objects, especially YOLOv7, and YOLOv6R3s, as they miss Upper Taskbars which can be easily detected using classic computer vision techniques.

Drawers and Switches are the easiest elements to detect. Drawers, which are also known as sliding menus, are easy to detect as they are very large and occupy more than half of the image. Switches, however, are very small, yet they are usually placed closer to the middle and right side of the images, and there are not many variations between different UIs for switches, thus it is easier for a neural network to generalize to different types of switches. 

Checked Text Views are hard to detect as they contain some text with a checkbox, where the checkbox comes in various forms and shapes, and are smaller compared to switches. As they are typically unchecked, the inside and outside of them are usually the same color as the background, and only narrow edges make them visible, which becomes even narrower when the image is resized into a smaller image. 

Hence, neural networks tend to lose the checkboxes, and not detect them at all; therefore, the only section of Checked Text Views that is usually detected is the text section of them. The detection performance might improve if larger images are used as the input of the models, but this requires more complex and deeper networks so that the models generalize to new data, yet this makes the model slower to train, and use for inference.

In conclusion, Checked Text Views are harder and Drawers and Switches are easier to detect and classify for all models, regardless of the fact that all three are the least common labels in the dataset, as described in figure \ref{fig:ld}. YOLOv8s is more sensitive to elements with a larger width-to-height ratio and performs poorly when elements are too wide and short at the same time.

\subsection{Results - RQ3 Verification}
The motivation behind this RQ is that many developers and researchers tend to trust the results of studies done on general datasets due to a lack of comparisons among the models on GUI element detection datasets, thus we investigate whether doing so is logical or not. We also reveal that trusting the validation set results, and not looking at the test set results especially on YOLOs can cause problems in understanding which model is best. 

Similar models were tested on the MS COCO dataset \cite{coco} by Lie et al \cite{yolov6r3}, where they reported the AP@.5 on the validation set of the MS COCO dataset on various models, including YOLOv5R6.1s, YOLOv6R3s, YOLOv7, and YOLOv8s. Hence, we similarly reported the performance of these models on our validation set of the VINS dataset, with the exception of YOLOv5R6.1s where its newer version (YOLOv5R7s) is used. It is worth noting that comparing models based on their performance on the validation set is not very logical as it is used in the training process, yet most papers only contain the results on the validation set.

\begin{table}[h!]
    \centering
\begin{tabular}{ |p{1.9cm}||p{2.7cm}|p{2.7cm}|  }
 \hline
 Model & Li et al. \cite{yolov6r3} (MS COCO dataset \cite{coco}) & Ours (VINS dataset \cite{VINS})\\
 \hline
YOLOv5R6.1s & 56.8 & - \\
YOLOv5R7s & - & 94.7\\
YOLOv6R3s & 61.8 & 95.1\\
YOLOv7 & 69.7 & 94.0\\
YOLOv8s & 61.8 & 93.3\\
 \hline

\end{tabular}

    \caption{AP@.5 of the selected models on the validation set of the VINS dataset, and similar models on the validation set of MS COCO \cite{yolov6r3}.}
    \label{table:comp}
\end{table}

As it is described in table \ref{table:comp}, we see that YOLOv7 outperforms all the other common models on the MS COCO dataset, while on the VINS dataset, YOLOv6s outperforms others. 
Moreover, there is a huge performance difference between these models on the MS COCO dataset and on the VINS dataset. This is easily explained by the fact that the MS COCO dataset contains 80 classes of objects compared to 12 classes in the VINS dataset, and it contains over 200K labeled images, whereas the VINS dataset contains less than 5000 images.
Also, the amount of variation among the GUI images is lower as the background is usually a uniform color and most elements do not vary much in terms of appearance, the number of objects in a GUI image is higher, plus the distances between the objects in GUI are lower, and the objects tend more to be in a rectangular shape or close to it in a GUI image, when compared to a natural and general dataset, such as MS COCO, where objects tend to have many degrees of freedom to change. Hence, we conclude that the results of these models on general datasets such as MS COCO do not necessarily apply to a use-case-specific dataset, especially on GUI element detection datasets such as VINS.

Another interesting fact is that in terms of AP@.5, the accuracy performance order between the models changes, as on the validation set YOLOv6R3s outperforms others, and on the test set YOLOv5R7s. This reveals that YOLOv6 heavily relies on the validation set for hyperparameter tuning, and more than others. For this reason, developers and researchers should not only rely on the validation set performance for the YOLO models, which is common, as they use the validation set to do automatic hyperparameter tuning, and the performance on the test set and the real-world unseen data might be different.

\section{Related Work}
GUI element detection is the first and most crucial stage of various software engineering tasks that deal with GUIs, such as GUI search \cite{Rico, VINS}, Code Generation \cite{UID2GUI2018, CodeGenEncDec2022, MLPrototypeMobile2020, REMAUI, image2emmet, xianyu}, UI Design Generation \cite{GUIGAN}, GUI Testing and Automation \cite{GuiTestWidget2019, AutoReportMobileDesignViolation2018}, and Advanced GUI interaction support \cite{WakenSV}. 

Chen et al. \cite{UIEDprior2020} was the first work that did a comprehensive comparison of methods for GUI element detection. They compared CenterNet \cite{Centernet}, YOLOv3 \cite{YOLO3}, Faster R-CNN \cite{FasterRCNN}, with two other methods from previous studies, namely REMAUI's method\cite{REMAUI}, Xianyu \cite{xianyu}, as well as their own proposed method that detects texts separately using an open source text detector and used old-fashioned computer vision algorithms for detecting non-text GUI elements. Xie et al. \cite{UIED} created a toolkit that contains the implementation of these methods and made all of these methods available on GitHub. Altinbas et al. \cite{altinbasYOLO5} used an older version of YOLOv5s \cite{yolov5} and trained it for 100 epochs in batches of 16 on the VINS dataset and reported their results, without specifying which version of YOLOv5 they used, and as our study already includes the latest version of YOLOv5 and it outperformed their work in all aspects we did not include their results in the tables.

Bunian et al. \cite{VINS} collected a dataset of Android, IOS, and abstract wireframes, used the dataset without the wireframes to train a GUI element detection model based on Singleshot Multibox Detector (SSD), and used the trained model to generate tentative segmented layout representations from the images and then train a model on them to learn a joint feature representation to retrieve similar UIs.

Zhao et al. \cite{GUIGAN} trains a Generative Adversarial Network (GAN) to generate new UI designs from a dataset of UI designs for designers. Their model sees the screenshots of an application and generates new creative UI designs.

Xu et al. \cite{image2emmet} collected images of webpages with their HTML-CSS code, then used region-based object detection deep learning models, and finally generated corresponding code for the components via a deep learning model, combining a CNN and an LSTM. Chen et al. \cite{CodeGenEncDec2022} similarly used a dataset to train an encoder-decoder network with an attention mechanism to generate code from UI images. Chen et al. \cite{UID2GUI2018} built a complex pipeline to acquire GUIs and transform them into GUI skeletons, using deep learning. Xianyu \cite{xianyu} is a code generation tool developed by Alibaba, which uses classic computer vision techniques for GUI element detection. REMAUI \cite{REMAUI} detects text and not-text GUI elements using an OCR tool and classic computer vision algorithms respectively. Moran et al. \cite{MLPrototypeMobile2020} proposed a system for generating code from images; they similarly leveraged classic computer vision algorithms for GUI element detection. Zhu et al. \cite{IntentGen} generate intents for GUI widgets using an encoder-decoder network for the intent and a pre-trained network for feature extraction from the images, as the initial step.

White et al. \cite{GuiTestWidget2019} used YOLOv2 \cite{YOLOv2} to detect and classify GUI elements from Java Swing Applications for GUI testing. Moran et al. \cite{AutoReportMobileDesignViolation2018} presented an approach for reporting design violations in mobile apps, and used classic computer vision algorithms for the detection and classification of elements.

Banovic et al. \cite{WakenSV} utilized a number of classic computer vision algorithms to extract usage information from videos without requiring any templates or specific rules.

\section{Limitations}
Here we highlight a few limitations of our work:
\begin{itemize}
    \item \textbf{Models' Variants:} We selected the S variants of YOLOv5, YOLOv6, and YOLOv8 to compare them with YOLOv7. We did not test the performance of the M variants, which are close to YOLOv7 in terms of the number of parameters. The reason we did not compare the M variants is that the S models already beat YOLOv7 in terms of AP@.5, thus it is expected that the M variants will beat it as well. Furthermore, the S variants are way faster than the M models and can be run on low-end systems with acceptable performance, hence it is likely that one chooses an S variant for GUI element detection, as dependent tasks such as Code Generation and GUI Testing require fast detection speeds.
    
    \item \textbf{Model Selection:} We only compared the results of the most recent YOLO models, yet there exist quite a lot more models that can be leveraged for GUI element detection, some of which might actually beat all of the selected models for GUI element detection, in terms of AP@.5 and other metrics.
    
    \item \textbf{Dataset} We only used the VINS dataset for training and evaluation of the selected models, which is a dataset that solely contains images of mobile applications, and might not apply completely to other datasets that target different environments such as web pages or desktop applications.
\end{itemize}

\section{Threats to Validity}
\begin{figure}[h!]
    \centering
    (a){\includegraphics[width=0.215\textwidth]{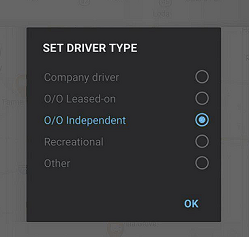}}
    (b){\includegraphics[width=0.215\textwidth]{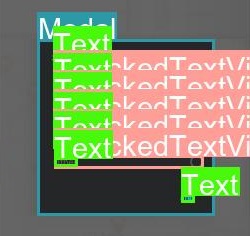}}
    \caption{A Cropped Sample Image and its labels (a) Cropped Image (b) Cropped Image with Original Labels}
    \label{fig:modal}
\end{figure}

\begin{figure*}[h!]
    \centering
    (a){\includegraphics[width=0.31\textwidth]{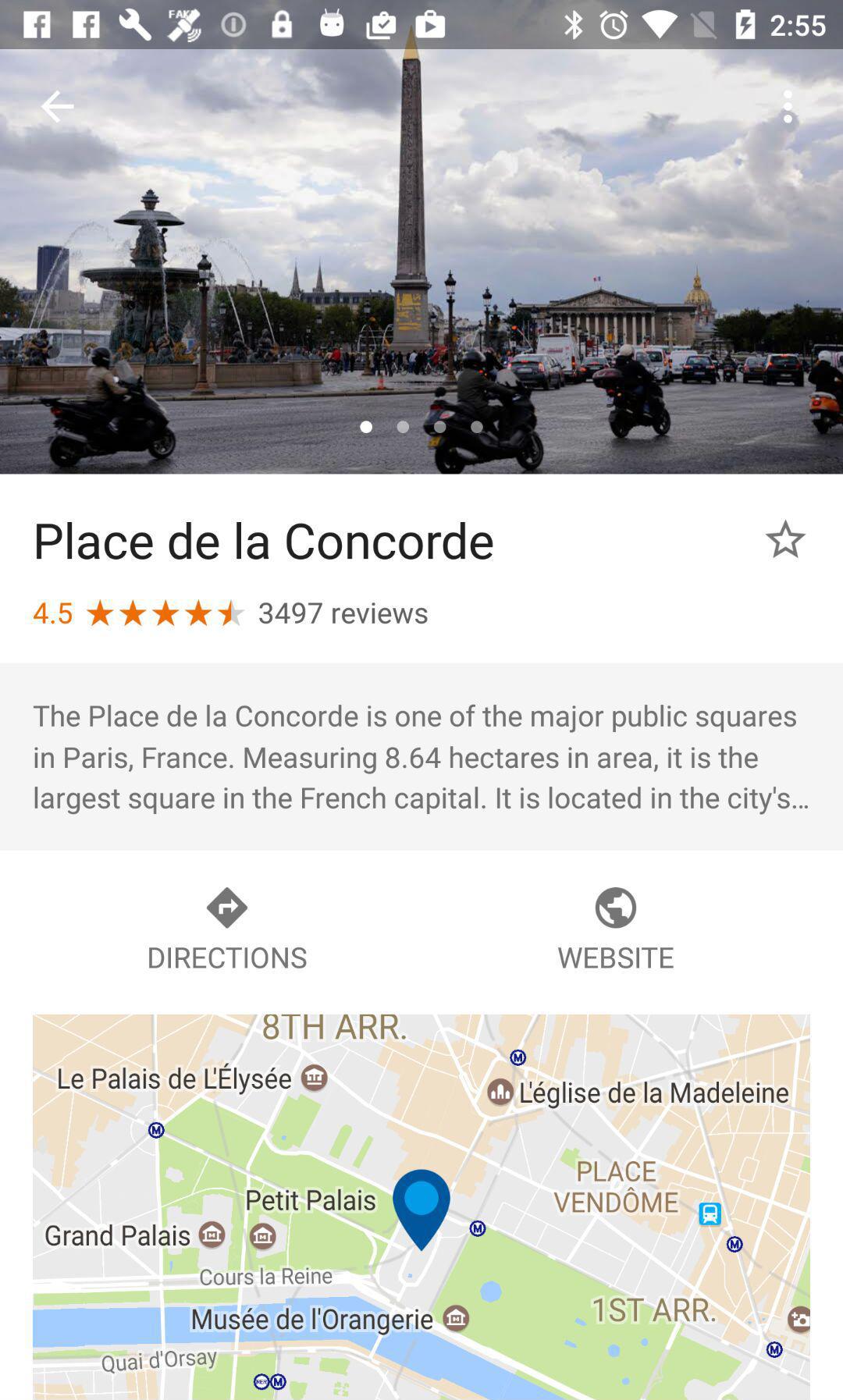}}
    (b){\includegraphics[width=0.31\textwidth]{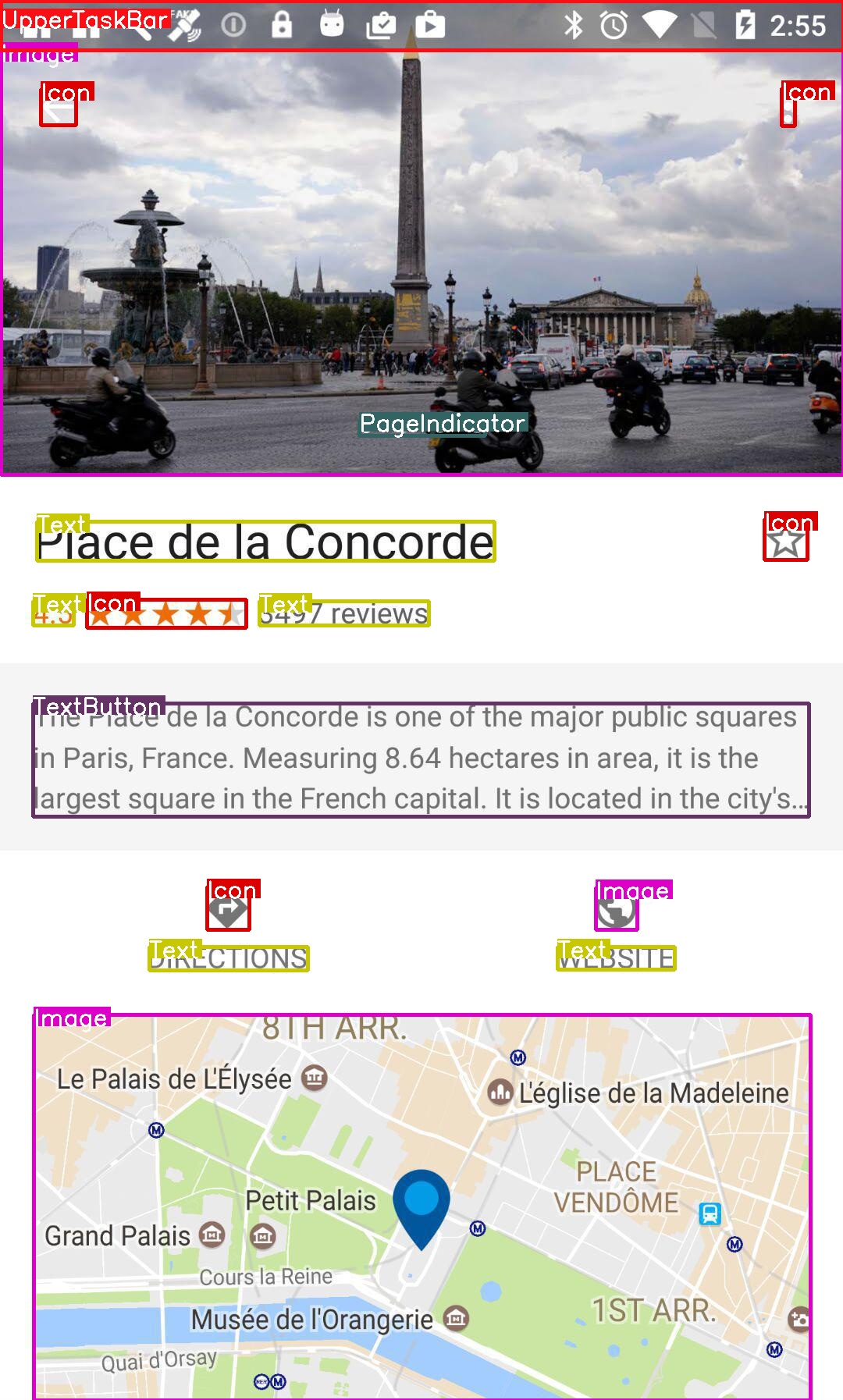}}
    (c){\includegraphics[width=0.31\textwidth]{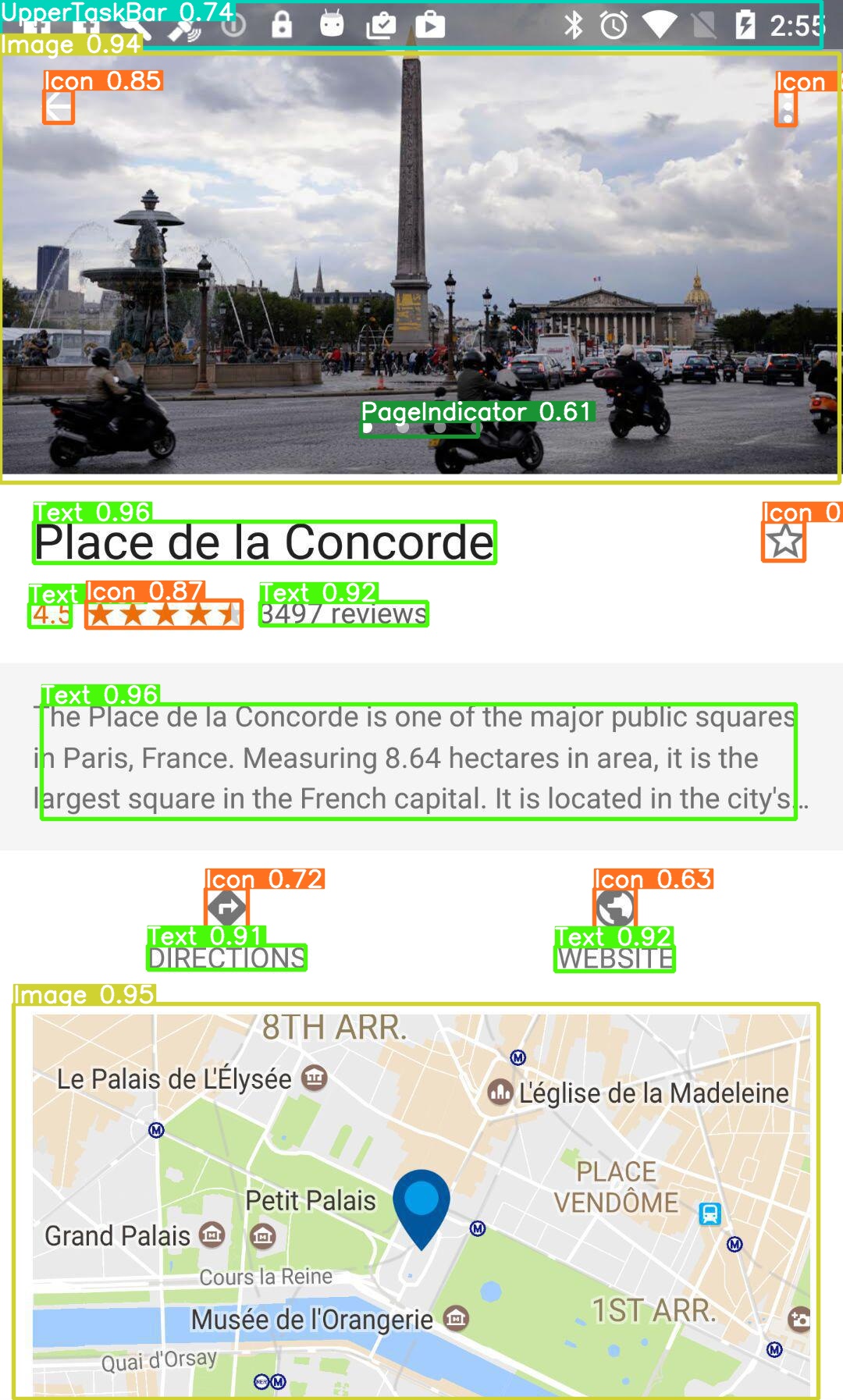}}
    (d){\includegraphics[width=0.31\textwidth]{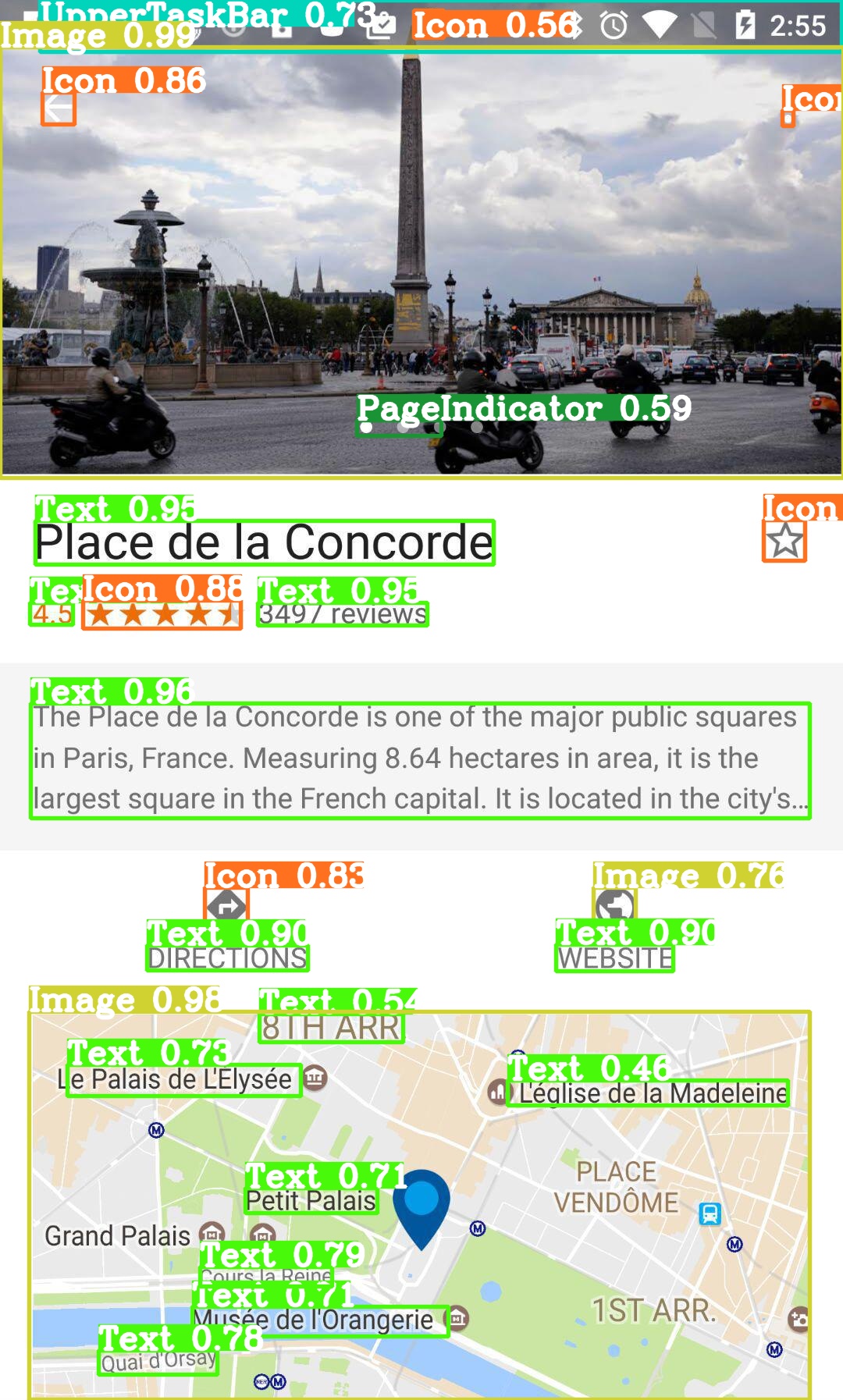}}
    (e){\includegraphics[width=0.31\textwidth]{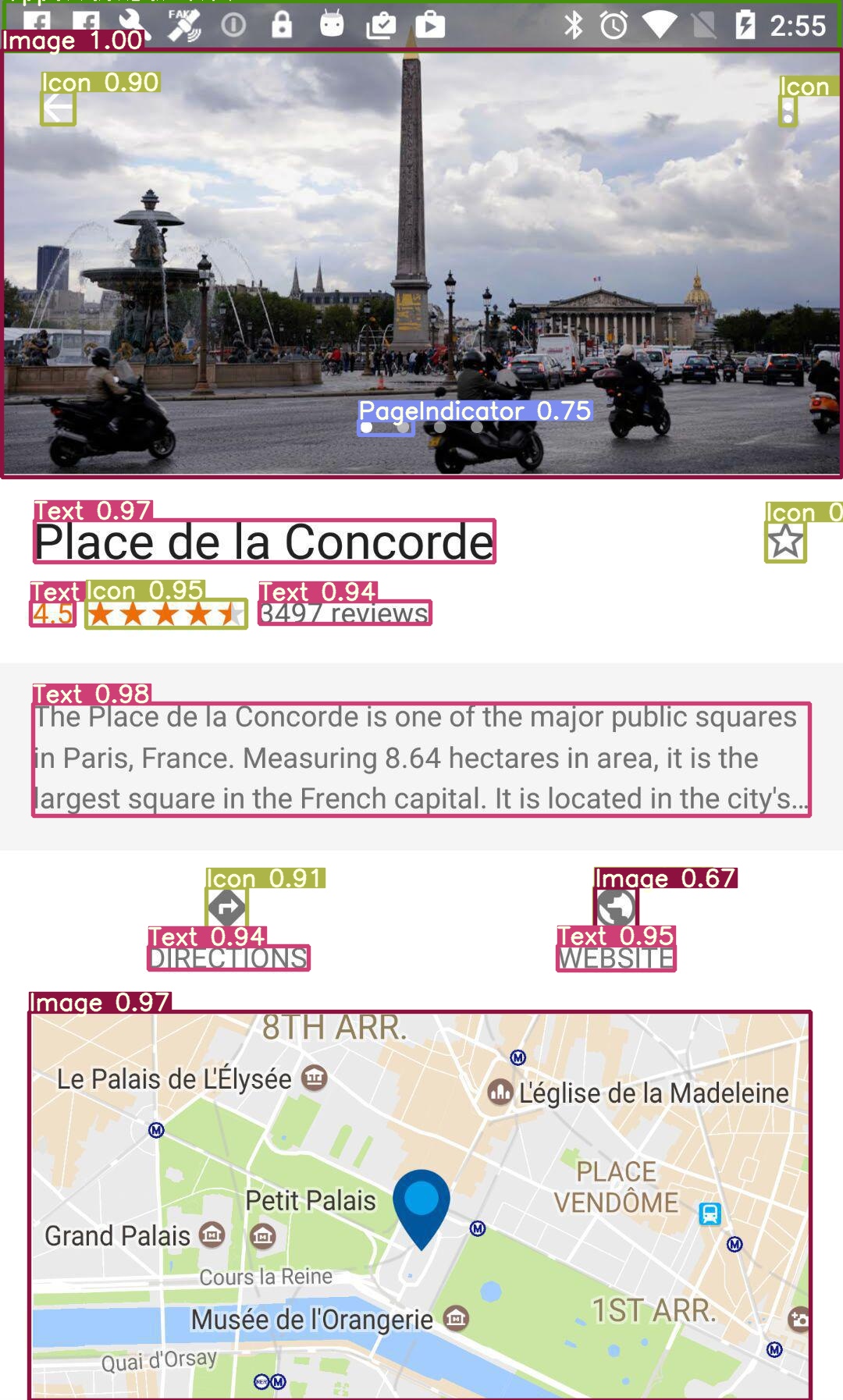}}
    (f){\includegraphics[width=0.31\textwidth]{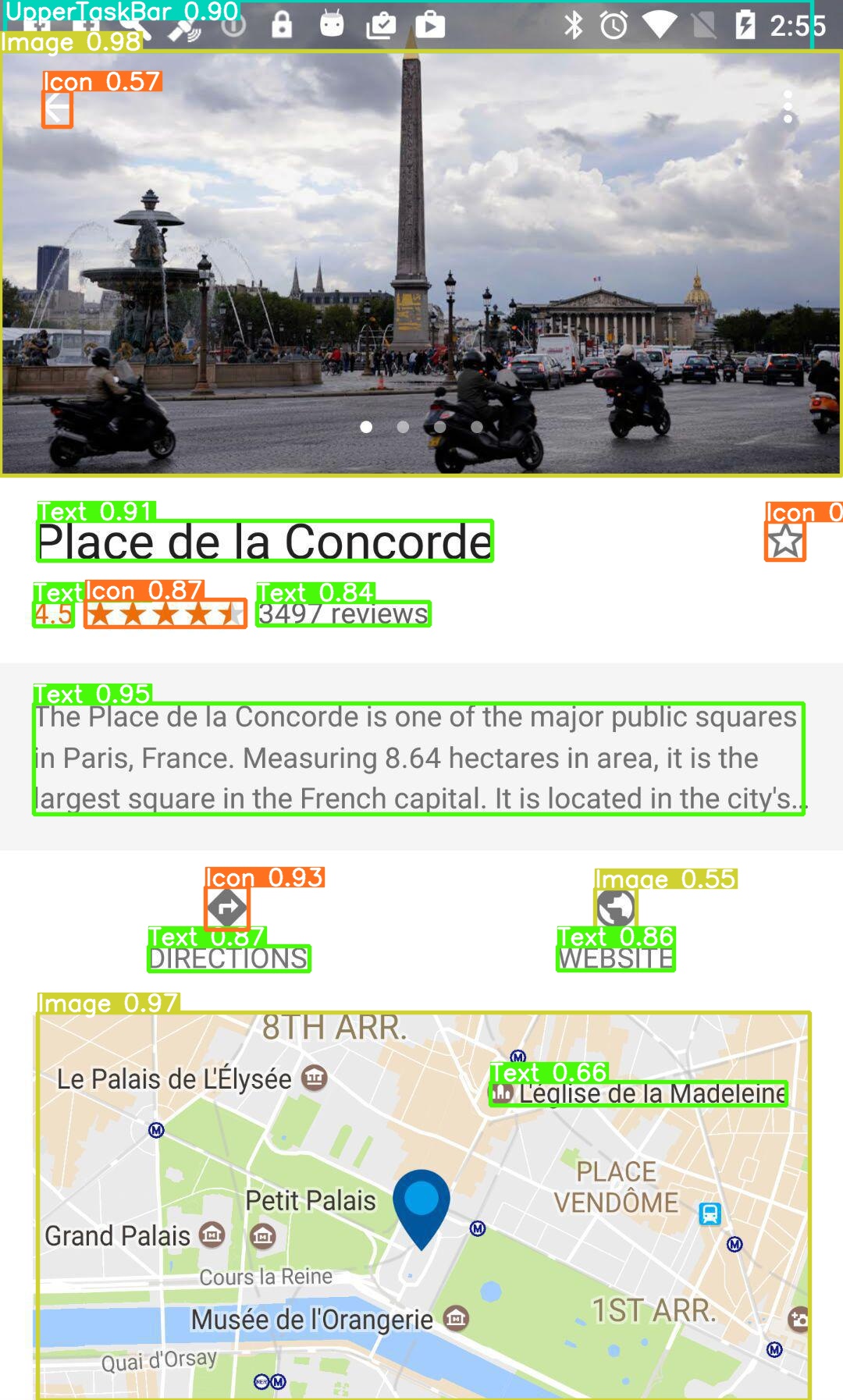}}
    \caption{Models' results on a test sample. (a) Original Image (b) Ground-Truth (c) YOLOv5R7s (d) YOLOv6R3s (e) YOLOv7, (f) YOLOv8s}
    \label{fig:allresults}
\end{figure*}

As in any study, some threats exist to the validity of the presented results. In the following we have identified and listed some of these threats:

\begin{itemize}
    \item \textbf{Label Noise:} It is possible that some of the labels in the dataset were labeled the wrong label or not labeled at all. Figure \ref{fig:modal} shows a sample where a Text Button is wrongly labeled as a Text (the OK button) in the dataset. This certainly affects the training and the evaluation of the models on the dataset. It is worth noting that some objects can be labeled as two things, e.g. in \ref{fig:allresults} there is an icon that is created using an image of a globe, and not using font-based icon toolkits such as Font Awesome. In such cases, the models might predict either of the labels, but only the one mentioned in the dataset will be considered the true label.
    \item \textbf{Model Bias:} Although we have mentioned the exact version of the models that we used, some of them such as YOLOv5 and YOLOv8 are being updated constantly on Github, without increasing their major version. This can change the achieved results to some extent if a similar study is carried out on the same models, with similar major versions but different minor versions.
    \item \textbf{Data Bias:} As with every study, there exists a data bias as we randomly sampled and separated the three sets of train, validation, and test sets. Although the selected models use loss functions that are designed to deal with class imbalance (such as Focal Loss \cite{focal_loss}) during training, the validation and test set might have had considerable differences, and as the validation set is used for hyperparameter tuning, the final models might do worse than intended on the test set compared to the validation set.
\end{itemize}

\section{Conclusion and Future Work}
This paper investigates the performance of the latest YOLO object detection deep learning models, namely YOLOv5R7s, YOLOv6R3s, YOLOv7, and YOLOv8. All of the models are trained on the VINS dataset, which contains Mobile GUI Images, with the goal of detecting and classifying GUI Elements. The results demonstrate that YOLOv5R7s achieves better overall performance in terms of AP@.5; Two classes of elements, namely drawers (also known as sliding menus) and switches are easier, and checked text views are the hardest to detect and classify. 

We also conclude that YOLOv8 performs poorly on wide but short elements, and YOLOv7 and YOLOv6 fail to detect and classify upper taskbars with great accuracy, while others do not. We, also show that the results of previous studies on comparing different YOLO models on general datasets do not necessarily apply to GUI element detection datasets due to their unique characteristics. Finally, we highlight the limitations of our work as well as possible threats to the validity of the presented results.

For future work, we plan to implement, train, and test different models for code generation with YOLOv5R7s and some classic computer vision techniques, and compare them with other end-to-end deep learning-based methods that transform images into code directly. We also plan to investigate the GUI element detection differences between mobile and web pages and see whether it is possible to build a code generation system that can generate code for the web and mobile without the use of cross-platform tools and frameworks, such as React Native, where a single source code is required to create applications for different platforms.

\bibliographystyle{ACM-Reference-Format}
\bibliography{sample-authordraft}

\appendix

\end{document}